\documentclass[11pt]{article}

\usepackage[preprint]{neurips_2026}   

\usepackage[utf8]{inputenc}
\usepackage[T1]{fontenc}
\usepackage{microtype}

\usepackage{amsmath, amssymb, amsthm, mathtools}
\usepackage{amsfonts}
\usepackage{bm}
\usepackage{nicefrac}

\usepackage{graphicx}
\usepackage{booktabs}
\usepackage{multirow}
\usepackage{subcaption}
\usepackage{float}

\usepackage{setspace}
\usepackage{enumitem}
\usepackage{xcolor}

\usepackage{algorithm}
\usepackage{algpseudocode}

\usepackage{listings}
\usepackage{hyperref}
\usepackage{url}

\providecommand{\safelambda}{\lambda}
\providecommand{\tspbudget}{N}
\providecommand{\tsprate}{\rho}
\providecommand{\tsplayer}{\ell^{*}}
\providecommand{\harmscore}{h}
\providecommand{\harmscorepos}{h^{+}}
\providecommand{\attnscore}{a}
\providecommand{\meanattn}{\bar{a}}
\providecommand{\anchor}{\bm{v}_{h}}
\providecommand{\windowsize}{w}
\providecommand{\promptlen}{L}
\providecommand{\score}{\mathrm{score}}
\providecommand{\TSPset}{\mathcal{S}}
\providecommand{\TopK}{\mathrm{TopK}}
\providecommand{\harmthresh}{\theta}

\theoremstyle{plain}

\theoremstyle{definition}

\theoremstyle{remark}

\title{\textbf{AnchorKV: Safety-Aware KV Cache Compression via\\
Soft Penalty with a Refusal Anchor}}

\author{%
  Ning Ni \\
  Department of Computer Science\\
  Tufts University\\
  \texttt{Ning.Ni@tufts.edu} \\
  \And
  Yingjie Lao \\
  Department of Electrical and Computer Engineering\\
  Tufts University \\
  \texttt{yingjie.lao@tufts.edu} \\
}

\begin{document}
\maketitle

\begin{abstract}
Large language models (LLMs) outperform earlier architectures on generative inference and long-context tasks, but their unprecedented size makes memory usage, energy cost, and on-device deployment a serious challenge. Because scaling pre-trained language models tends to improve downstream capacity \cite{zhao2023survey}, the resulting key-value (KV) cache becomes a dominant inference cost. Recent KV cache compression methods \cite{jo2025fastkv,li2024snapkv,zhou2024dynamickv} reduce this cost by retaining only a small subset of attention-relevant tokens. Although these methods preserve task accuracy on benign workloads, their compression policies either fail to defend against jailbreak attacks \cite{jiang2024robustkv} or actively erode safety alignment under aggressive token eviction. We propose AnchorKV, a drop-in modification to KV cache compression that biases the per-token retention score against directions in key space associated with harmful prompts. AnchorKV uses an offline anchor construction procedure that adapts the difference-of-means representation-engineering technique \cite{arditi2024refusal,zou2023representation} to the layer-specific key projection space relevant for KV cache compression. Given this anchor, a soft-penalty token-selection rule trades a small amount of utility for a substantial improvement in safety alignment, while reducing bit-exactly to the unmodified compressor at zero penalty.
\end{abstract}

\section{Introduction}
\label{sec:intro}
Long-context language models have rapidly become a central deployment
target for large language models (LLMs), with context windows expanding
from 4K to 128K tokens and beyond.
At inference time, this scaling shifts the dominant cost from parameter
count to the key--value (KV) cache, whose memory and latency grow
linearly in sequence length.
Modern KV cache compression methods reduce this cost by retaining only
a small subset of attention-relevant tokens, and have been shown to
preserve task accuracy on standard long-context
benchmarks~\cite{li2024snapkv,jo2025fastkv,zhou2024dynamickv}.
 
However, the accuracy-centric design of these methods leaves their
interaction with the model's safety alignment largely unexamined.
Recent work~\cite{jiang2024robustkv} reports that off-the-shelf KV
compression offers essentially no defense against jailbreak attacks:
applying SnapKV~\cite{li2024snapkv} at typical compression rates barely
changes the attack success rate of GCG-~\cite{zou2023universal} or
AutoDAN-style~\cite{liu2023autodan} prompts.
Our own preliminary experiments additionally confirm a related but
under-discussed pattern: aggressive token eviction can not only
fail to mitigate jailbreaks, but actively distort the model's response
distribution in safety-relevant ways.

\begin{figure}[t]
  \centering
  \includegraphics[width=0.95\linewidth]{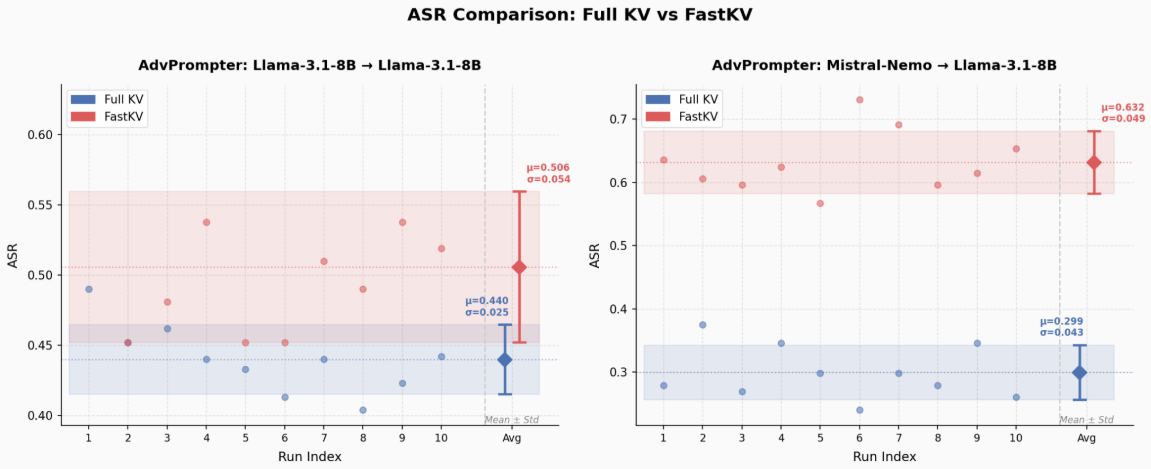}
  \caption{
    Compression alone increases jailbreak vulnerability.
    Attack Success Rate (ASR) of AdvPrompter
    on Llama-3.1-8B-Instruct~\cite{grattafiori2024llama} under FullKV (no
    compression) and FastKV ($\tsprate=0.25$),
    measured across $10$ random seeds.
    \textbf{Left}: white-box self-attack with Llama-3.1-8B-Instruct as
    both attacker network and target.
    FastKV raises mean ASR from $0.440$ to $0.506$
    (a $15\%$ relative increase).
    \textbf{Right}: transfer attack from Mistral-NeMo-Instruct-2407
    \cite{mistralnemo2024} to the same Llama-3.1-8B-Instruct target.
    The effect is more dramatic: FastKV more than doubles mean ASR
    from $0.299$ to $0.632$, with the two distributions barely
    overlapping.
    Diamonds with error bars mark the across-seed mean and standard
    deviation;
    individual dots are per-seed scores.
    The gap is consistent across both protocols, indicating a property
    of the compression policy rather than of any specific attacker.
  }
  \label{fig:motivation_asr}
\end{figure}
 
We begin with a direct empirical observation
(Figure~\ref{fig:motivation_asr}): on the same Llama-3.1-8B-Instruct
target, simply replacing FullKV with FastKV at a $25\%$ retention
rate raises the AdvPrompter~\cite{paulus2024advprompter} jailbreak
ASR from $0.440$ to $0.506$ under self-attack, and from $0.299$
to $0.632$ under a Mistral-NeMo cross-model transfer attack---a
relative increase of $15\%$ in the former and over $100\%$ in the
latter.
The effect is consistent across $10$ random seeds and across both
attack protocols, indicating a structural rather than incidental
phenomenon.
We characterize two distinct pathologies underlying it.
\textbf{(i)~Refusal-evidence dilution}: when a jailbreak prompt's
adversarial framing accumulates high attention scores, the harmful
query tokens become eviction candidates under any attention-greedy
policy; with enough of them dropped, the downstream layers no longer
see sufficient evidence to trigger refusal, and the model complies
with attacks it would have rejected at full capacity.
The widening gap under cross-model transfer attack
(Figure~\ref{fig:motivation_asr}, right) supports this reading:
prompts whose harmful content is distributed across more positions
(a typical signature of cross-model transferability) are
disproportionately disrupted by attention-greedy eviction.
\textbf{(ii)~Template-mode regression}: conversely, when
prompt-content tokens are dropped while harm-related skeleton tokens
(generic verbs, help-seeking phrases) are retained, the model
regresses to formulaic safety templates---enumerating mental-health
resources or generic ``I cannot assist'' boilerplate---that bear no
relation to the user's actual query.
\emph{Both pathologies are artifacts of the compression policy}
rather than properties of the underlying model, and both grow with
the compression ratio.

Prior work on \emph{RobustKV}~\cite{jiang2024robustkv} successfully addresses jailbreaks by evicting low-attention tokens at every head, but does so as a stand-alone defense: it neither compresses the cache nor composes
with existing accuracy-oriented compressors, and its discriminative
signal---attention rank---is precisely the quantity that an adaptive
adversary can manipulate.
Our goal is fundamentally different: \emph{to preserve safety alignment
under aggressive KV compression}.
This requires a safety signal that is independent of attention ranking
and that operates at the same operational point (the eviction layer)
as the compressor itself.
We argue that recent representation-engineering
work~\cite{zou2023representation,arditi2024refusal} suggests such a
signal: a single linear ``refusal direction'' in the model's internal
representations, recoverable from a small set of contrastive prompts.
Section~\ref{sec:background} reviews these threads in detail and makes
the contrast precise.
 
We introduce \emph{AnchorKV}, a drop-in scalable modification to KV cache
compression in which the per-token retention score is augmented
with a soft penalty derived from a precomputed harm anchor.
Concretely, we offline construct a single direction
$\bm{v}_{h} \in \mathbb{R}^{H \times d}$ in the layer-$\ell^{*}$
\emph{key projection space}---rather than the residual stream---by
applying the difference-of-means recipe of~\cite{arditi2024refusal} to
length-matched harmful and harmless prompts from the AdvPrompter
60/20/20 split of AdvBench~\cite{paulus2024advprompter,zou2023universal}.
At inference, every prompt token receives a harm score equal to its
projection onto~$\bm{v}_{h}$; this score is converted into a soft
penalty subtracted from the attention importance score with strength
$\lambda$, named as safe rate.
Three design properties distinguish AnchorKV from prior work.
\textbf{(a)~Orthogonal signal}: the harm score is computed from the
geometry of the key space, not from attention ranking, and is
therefore not subject to the evasiveness dilemma of attention-based
defenses.
\textbf{(b)~Continuous and Scalable reduction to baseline}: the penalty is
continuous in $\lambda$ and reduces \emph{bit-exactly}
to the unmodified FastKV baseline at $\lambda = 0$, enabling rigorous ablation. The penalty of harm scores can be applied on other Key-level compression techniques by scaling the safe rate $\lambda$.
\textbf{(c)~Layer alignment with eviction}: because the anchor is
constructed in the same layer at which token eviction occurs, the
geometric calibration matches the operational point---a constraint
that residual-stream-based methods cannot satisfy without an
additional projection step.
 
\paragraph{Contributions.} We make four contributions:
\begin{enumerate}[leftmargin=1.25em,topsep=2pt,itemsep=2pt]
    \item We empirically characterize a previously under-documented
    interaction between KV cache compression and safety alignment,
    by observing a regression of safety alignment that arises from
    token eviction at high compression ratios.
 
    \item We adapt the difference-of-means representation-engineering
    technique~\cite{zou2023representation,arditi2024refusal} from the
    residual stream to the per-head key-projection space of a
    designated eviction layer.
    The resulting anchor achieves a held-out test AUROC of 0.996 on harmful
    versus harmless discrimination on the AdvPrompter held-out test
    split, validating that the linear refusal direction observed in
    residual activations also exists, and is recoverable, in
    layer-specific key space.
 
    \item We design a soft-penalty token-selection rule that is
    continuous in its strength parameter $\lambda$
    and reduces bit-exactly to the FastKV baseline at $\lambda = 0$.
    This property allows AnchorKV to be ablated as a strict superset
    of FastKV, and any other similar token eviction techniques by scaling $\lambda$.
 
    \item We report the safety--utility trade-off frontier of
    AnchorKV on jailbreak benchmarks and
    LongBench~\cite{bai2024longbench}, identifying a non-monotonic
    \emph{reversal regime} in which excessive penalty strength
    cannibalizes the prompt's content backbone.
    This finding motivates the structural safeguards in our method
    and offers a cautionary lesson for future safety-aware compression
    work: more safety pressure does not monotonically yield safer
    behavior.
\end{enumerate}
 
\section{Background and Related Work}
\label{sec:background}
 
We survey three threads relevant to AnchorKV: KV cache compression
(§\ref{sec:bg_kv}), jailbreak attacks and defenses on models after KV cache compression (§\ref{sec:bg_jail})
 and representation engineering (§\ref{sec:bg_repe}).
The synthesis at the end of §\ref{sec:bg_repe} states the gap our
work fills.
 
\subsection{KV Cache Compression}
\label{sec:bg_kv}
 
The KV cache stores per-layer key and value tensors of all preceding
tokens, enabling auto-regressive decoding without recomputing the
prefix.
Its size grows linearly in sequence length, becoming the dominant
memory bottleneck for long-context inference.
 
\paragraph{Eviction-based methods.}
H2O~\cite{zhang2023h2o} introduces the ``heavy-hitter'' principle:
during decoding, tokens that have historically received high attention
are retained while others are greedily evicted.
Scissorhands~\cite{liu2023scissorhands} extends this with a notion of
attention persistence across decoding steps, while
StreamingLLM~\cite{xiao2023efficient} highlights the role of the
first few tokens (``attention sinks''), whose aberrant key statistics
make them indispensable regardless of semantic content.
SnapKV~\cite{li2024snapkv} adapts these ideas to the prefilling stage:
for each attention head, an observation window of recent tokens
identifies which earlier positions to keep, and eviction is performed
independently per head.
DynamicKV~\cite{zhou2024dynamickv} further makes the retention budget
input-adaptive.
FastKV~\cite{jo2025fastkv} introduces \emph{token-selective
propagation} (TSP): up to a designated layer~$\ell^{*}$, all tokens
participate in attention; beyond~$\ell^{*}$, only a fraction~$\rho$
of tokens---selected by an attention-cache importance score---is
propagated to the remaining layers.
This last design is the closest to ours: we adopt FastKV as our
backbone, leaving the per-layer KV compression untouched and modifying
only the TSP scoring rule at layer~$\ell^{*}$. Other methods that select tokens via an additive importance score can also be combined with AnchorKV.
 
\subsection{Jailbreak Attacks and Defenses}
\label{sec:bg_jail}
 
\paragraph{Attacks.}
Jailbreak attacks~\cite{wei2023jailbroken} embed a harmful query
within a wrapper prompt designed to bypass the model's built-in
safeguards.
Optimization-based attacks such as GCG~\cite{zou2023universal} append
adversarial suffixes that maximize the model's probability of an
affirmative response;
AutoDAN~\cite{liu2023autodan} performs genetic search over a population
of handcrafted templates;
AdvPrompter~\cite{paulus2024advprompter} trains an auxiliary LLM to
generate fluent adversarial prompts.
For all such attacks, the standard benchmark is
AdvBench~\cite{zou2023universal}, with the 60/20/20 train/validation/test
split introduced by AdvPrompter.
 
\paragraph{Prompt-level defenses.}
SmoothLLM~\cite{robey2023smoothllm} applies randomized character-level
perturbations to disrupt brittle adversarial suffixes;
GoalPriority~\cite{zhang2024defending} crafts a system prompt that
instructs the model to prioritize safety over helpfulness.
Both operate at the prompt surface level and do not address how the
model's internal computation handles harmful content.
 
\paragraph{KV-cache-level defenses.}
The most relevant prior work is \emph{RobustKV}~\cite{jiang2024robustkv}.
It observes empirically that, for a jailbreak prompt to bypass the
LLM's safeguards, its adversarial framing must accumulate enough
attention to outrank the harmful query tokens; consequently, the
harmful query tokens tend to be among the lowest-ranked positions by
attention-cache score.
RobustKV exploits this pattern by evicting the lowest-ranked $p\%$ of
tokens (typically~$20\%$) at every attention head, suppressing the
harmful query's representation in the cache.
Although effective as a defense, RobustKV diverges from our setting in
three respects.
First, it is a \emph{defense}, not a compressor: it replaces $p\%$ of
cache positions with low-attention tokens but reports no memory
savings, leaving open whether the mechanism survives \emph{on top of}
aggressive compression ratios such as those targeted by FastKV.
Second, its discriminative signal is attention rank itself, which an
adaptive adversary can manipulate; the authors identify an
``evasiveness dilemma'' in which the attacker trades attention rank
against safeguard bypass.
Third, the assumption that ``harmful query tokens are always
low-attention'' is empirical rather than mechanistic, and is not
guaranteed for jailbreak families that elevate the harmful query's
attention via role-playing scaffolds.
We are not aware of prior work that pursues safety preservation
\emph{under} aggressive KV compression, with a discriminative signal
independent of attention ranking.
 
\paragraph{Safety under model compression.}
Beyond KV compression specifically, recent work~\cite{hong2024decoding}
has begun to scrutinize how parameter quantization and pruning affect
the trustworthiness of LLMs, generally finding that aggressive
compression can erode safety properties even when accuracy is
preserved.
Our findings reinforce this concern within the orthogonal axis of KV
cache compression.
 
\subsection{Representation Engineering and the Refusal Direction}
\label{sec:bg_repe}
 
A growing body of work studies whether high-level concepts in LLMs are
encoded as linear directions in the model's internal activations.
Zou et al.~\cite{zou2023representation} introduce \emph{representation
engineering} (RepE) as a top-down framework for extracting and
manipulating such directions, demonstrating that diverse concepts
ranging from honesty to harm can be identified via contrastive prompt
pairs.
Arditi et al.~\cite{arditi2024refusal} specialize this analysis to
\emph{refusal} in instruction-tuned LLMs and show, on a wide range of
models, that refusal is mediated by a \emph{single} linear direction
in the residual stream, extractable as the difference-of-means of
activations between harmful and harmless prompts.
Removing this direction from activations bypasses refusal; conversely,
adding it strengthens refusal.
 
These results are obtained on residual-stream activations, which is a
natural choice for studies of model behavior at the granularity of
generation but mismatches the operational point of KV cache
compression: KV compressors operate on the per-head key projection
$K^{(\ell, h)}$ at each layer, not on the residual stream.
Naively transferring a residual-stream direction into the KV
compression decision incurs an additional projection step and a
geometric mismatch.
We instead construct the refusal direction \emph{directly in the
layer-$\ell^{*}$ key projection space}, ensuring that the safety
signal lives in the same vector space as the quantities being scored
for retention.
Whether the linear refusal direction survives this transfer is an
empirical question we answer in
Section~\ref{sec:exp}: it does, with a held-out test AUROC of 0.996 on
held-out test prompts.
 
\paragraph{Synthesis.}
We synthesize the three threads as follows.
KV compression methods (§\ref{sec:bg_kv}) reduce inference cost but
have not been designed with safety in mind; jailbreak defenses
(§\ref{sec:bg_jail}) preserve safety but either operate at the prompt
surface or impose a stand-alone KV manipulation incompatible with
aggressive compression; representation engineering
(§\ref{sec:bg_repe}) provides the mechanistic substrate for a
compression-compatible safety signal but has not been deployed in this
context.
AnchorKV bridges the three: it adds a RepE-derived signal to a
state-of-the-art compressor (FastKV), in the same vector space at the
same operational point, with a continuous strength
parameter that admits clean ablation against the unmodified baseline.
 
\section{Method}
\label{sec:method}
AnchorKV modifies \emph{only} the TSP scoring rule of
FastKV at the designated layer~$\tsplayer$;
the per-layer KV compression and the local-window mechanism are
inherited unchanged.
We use the FastKV-defined quantities introduced in §\ref{sec:bg_kv}
without re-derivation: the per-token importance score $\attnscore_t$,
the retention budget $\tspbudget = \lfloor \tsprate \cdot \promptlen
\rfloor$, the window size $\windowsize$, and the activation condition
$\promptlen > \tspbudget > \windowsize$.
 
The method is split into two phases.
The offline phase (§\ref{sec:method_offline}) constructs a single
direction $\anchor \in \mathbb{R}^{H \times d}$ in the
layer-$\tsplayer$ key projection space, plus a calibrated threshold
$\harmthresh$, from a small set of contrastive prompts.
The online phase (§\ref{sec:method_online}) consumes $(\anchor,
\harmthresh)$ at every TSP-triggering prefill, augmenting
$\attnscore_t$ with a soft penalty before top-$k$ selection.
 
\subsection{Offline Anchor Construction}
\label{sec:method_offline}
 
\paragraph{Difference of class means in key space.}
We adapt the difference-of-means recipe of
Arditi et al.~\cite{arditi2024refusal} from residual-stream activations
to the per-head key projection at layer $\tsplayer$.
Concretely, given two prompt sets---harmful $\mathcal{D}_h^{\mathrm{tr}}$
drawn from the AdvBench training
split~\cite{paulus2024advprompter,zou2023universal} and harmless
$\mathcal{D}_s^{\mathrm{tr}}$ drawn from a length-matched sample that is generated by Deepseek-V3 \cite{liu2024deepseek}---we run each prompt $p$
through the frozen model with a forward hook on the layer-$\tsplayer$
$K$-projection, obtaining
$K^{(\tsplayer)}(p) \in \mathbb{R}^{H \times L_p \times d}$, where $H$
is the number of key-value heads, $L_p$ the prompt length, and $d$ the
per-head dimension.
We summarize each prompt by all-token mean pooling
\begin{equation}
  \phi(p)
  \;=\;
  \frac{1}{L_p}\sum_{t=1}^{L_p} K_t^{(\tsplayer)}(p)
  \;\in\; \mathbb{R}^{H \times d},
  \label{eq:phi}
\end{equation}
and define the harm anchor as the difference of class means:
\begin{equation}
  \boxed{\;
  \anchor
  \;=\;
  \frac{1}{|\mathcal{D}_h^{\mathrm{tr}}|}
       \sum_{p \in \mathcal{D}_h^{\mathrm{tr}}} \phi(p)
  \;-\;
  \frac{1}{|\mathcal{D}_s^{\mathrm{tr}}|}
       \sum_{p \in \mathcal{D}_s^{\mathrm{tr}}} \phi(p).
  \;}
  \label{eq:anchor}
\end{equation}
Constructing the anchor in the same space at which TSP eviction
operates is critical: it eliminates the geometric mismatch that would
arise from transferring a residual-stream direction through a single
$K$-projection, and ensures that the harm signal lives in the same
coordinates as the quantities being scored at retention time.
 
\paragraph{Per-token harm score.}
At inference, each prompt token's harm tendency is the projection of
its layer-$\tsplayer$ key onto the unit-normalized anchor, averaged
across heads:
\begin{equation}
  \harmscore_t
  \;=\;
  \frac{1}{H}\sum_{h=1}^{H}
    \bigl\langle
      K_t^{(\tsplayer, h)},\;
      \widehat{\anchor}^{(h)}
    \bigr\rangle,
  \qquad
  \widehat{\anchor}^{(h)}
  \;=\;
  \anchor^{(h)} / \|\anchor^{(h)}\|_{2}.
  \label{eq:harm_score}
\end{equation}
 
\paragraph{Threshold calibration.}
We empirically observe that, on a frozen instruction-tuned LLM, both
class distributions of $\harmscore$ lie below zero, reflecting a shared bias direction in
the layer-$\tsplayer$ key space (e.g.\ attention-sink and
frequency components) that contributes a uniform negative offset to
all projections.
Using the projection sign as the harm/safe boundary therefore
systematically under-applies the penalty.
We calibrate an explicit threshold $\harmthresh$ on a held-out
validation split:
\begin{equation}
  \harmthresh
  \;=\;
  \tfrac{1}{2}\!\bigl(
    \mathrm{median}\{\harmscore(p)\!:\!p \in \mathcal{D}_h^{\mathrm{val}}\}
    +
    \mathrm{median}\{\harmscore(p)\!:\!p \in \mathcal{D}_s^{\mathrm{val}}\}
  \bigr),
  \label{eq:threshold}
\end{equation}
i.e.\ the midpoint between the two class medians, where
$\harmscore(p) = \frac{1}{L_p}\sum_t \harmscore_t$ is the prompt-level
mean of Eq.~\eqref{eq:harm_score}.
This places the discriminative boundary at the empirical crossing
point of the two classes; it does not affect the continuity property
of the online phase (§\ref{sec:method_online}).
 
\paragraph{Layer selection.}
The anchor is layer-specific by construction.
We select $\tsplayer$ on the validation split by sweeping a small set
of candidate layers, building the anchor of Eq.~\eqref{eq:anchor} at
each, and choosing the one that maximizes the AUROC of harm-vs.-safe
discrimination using prompt-level scores
$\{\harmscore(p)\}$.
 
\paragraph{Algorithm.}
Algorithm~\ref{alg:offline} summarizes the offline phase.
The output $(\anchor, \harmthresh, \tsplayer)$ is saved to disk and
loaded once at inference initialization;
no further forward passes through the frozen model are required at
runtime beyond the model's own.
 
\begin{algorithm}[t]
\caption{Offline anchor construction.}
\label{alg:offline}
\begin{algorithmic}[1]
\Require Frozen LLM $M$;
         training prompts $\mathcal{D}_h^{\mathrm{tr}},
         \mathcal{D}_s^{\mathrm{tr}}$;
         validation prompts $\mathcal{D}_h^{\mathrm{val}},
         \mathcal{D}_s^{\mathrm{val}}$;
         candidate layer set $\mathcal{L}$.
\Ensure Harm anchor $\anchor$, threshold $\harmthresh$,
        layer index $\tsplayer$.
 
\For{each candidate layer $\ell \in \mathcal{L}$}
    \For{each $p \in \mathcal{D}_h^{\mathrm{tr}} \cup
                       \mathcal{D}_s^{\mathrm{tr}}$}
        \State Run $M$ on $p$ with a hook on
               layer-$\ell$ $K$-projection;
               capture $K^{(\ell)}(p)$.
        \State Compute $\phi^{(\ell)}(p)$
               via Eq.~\eqref{eq:phi}.
    \EndFor
    \State $\anchor^{(\ell)}
           \gets
           \mathrm{mean}_{p \in \mathcal{D}_h^{\mathrm{tr}}}
              \phi^{(\ell)}(p)
           -
           \mathrm{mean}_{p \in \mathcal{D}_s^{\mathrm{tr}}}
              \phi^{(\ell)}(p)$
           \Comment{Eq.~\eqref{eq:anchor}}
    \State Score every $p \in \mathcal{D}_h^{\mathrm{val}} \cup
                                \mathcal{D}_s^{\mathrm{val}}$
           with Eq.~\eqref{eq:harm_score} using
           $\anchor^{(\ell)}$;
           record per-prompt mean $\harmscore^{(\ell)}(p)$.
    \State $\mathrm{AUROC}^{(\ell)}
           \gets$ AUROC of
           $\{\harmscore^{(\ell)}(p)\}$
           against the harm/safe label.
\EndFor
 
\State $\tsplayer
       \gets
       \arg\max_{\ell \in \mathcal{L}} \mathrm{AUROC}^{(\ell)}$
       \Comment{validation-only; test split not used}
\State $\anchor \gets \anchor^{(\tsplayer)}$
\State $\harmthresh
       \gets
       \frac{1}{2}\bigl(
         \mathrm{median}\{\harmscore^{(\tsplayer)}(p) :
                            p \in \mathcal{D}_h^{\mathrm{val}}\}
         +
         \mathrm{median}\{\harmscore^{(\tsplayer)}(p) :
                            p \in \mathcal{D}_s^{\mathrm{val}}\}
       \bigr)$
       \Comment{Eq.~\eqref{eq:threshold}}
\State \Return $(\anchor,\, \harmthresh,\, \tsplayer)$
\end{algorithmic}
\end{algorithm}
 
\subsection{Online TSP Selection with Soft Penalty}
\label{sec:method_online}
 
At every prefill that triggers the TSP branch
(see~§\ref{sec:bg_kv}), AnchorKV replaces the FastKV scoring rule
with
\begin{equation}
  \boxed{\;
  \score_t
  \;=\;
  \attnscore_t
  \;-\;
  \safelambda \cdot \meanattn \cdot \widetilde{\harmscorepos}_t,
  \;}
  \label{eq:safe_score}
\end{equation}
where $\safelambda$ is a non-negative penalty strength parameter,
$\meanattn = \frac{1}{\promptlen - \windowsize} \sum_t \attnscore_t$
is the mean of the FastKV importance score over non-window
positions, and $\widetilde{\harmscorepos}_t$ is the
\emph{safeguarded} harm intensity (defined below).
The TSP retention set is the standard top-$k$ on $\score_t$:
\begin{equation}
  \TSPset
  \;=\;
  \TopK_{t \in [0,\, \promptlen - \windowsize)}\!
       \bigl(\score_t,\; \tspbudget - \windowsize\bigr)
  \;\cup\;
  [\promptlen - \windowsize,\, \promptlen).
  \label{eq:select}
\end{equation}
 
\paragraph{Harm intensity with safeguards.}
Starting from the raw projection $\harmscore_t$ of
Eq.~\eqref{eq:harm_score} and the calibrated threshold $\harmthresh$
(Eq.~\eqref{eq:threshold}), we form the non-negative harm signal
\begin{equation}
  \harmscorepos_t \;=\; \max(\harmscore_t - \harmthresh,\; 0).
  \label{eq:harm_pos}
\end{equation}
We then apply two structural safeguards to obtain the final
$\widetilde{\harmscorepos}_t$.
\textbf{Sink protection}: positions $t < S$ are exempted from the
penalty, motivated by the anomalous key statistics of attention-sink
tokens~\cite{xiao2023efficient}, whose eviction triggers immediate
generation collapse:
\begin{equation}
  \widetilde{\harmscorepos}_t \leftarrow 0
  \quad \text{for } t < S
  \quad (\text{default } S = 4).
  \label{eq:sink}
\end{equation}
\textbf{Importance immunity}: the
$\lceil \alpha (\promptlen - \windowsize) \rceil$ positions with
the largest baseline scores $\attnscore_t$ are also exempted,
preventing the penalty from cannibalizing the prompt's content
backbone:
\begin{equation}
  \widetilde{\harmscorepos}_t \leftarrow 0
  \quad \text{for } t \in
  \TopK_{t}\!\bigl(
    \attnscore_t,\;
    \lceil \alpha (\promptlen - \windowsize) \rceil
  \bigr)
  \quad (\text{default } \alpha = 0.30).
  \label{eq:immune}
\end{equation}
Both safeguards act multiplicatively on the penalty (via masking
$\widetilde{\harmscorepos}_t$), preserving the continuity property
stated next.
 
\paragraph{Continuity at $\safelambda = 0$.}
A central design property of Eq.~\eqref{eq:safe_score} is that, when
$\safelambda = 0$, the penalty term is identically the zero tensor
\emph{regardless of} $\widetilde{\harmscorepos}_t$, $\meanattn$,
$\harmthresh$, or any safeguard mask:
\begin{equation}
  \TSPset^{(\safelambda=0)}
  \;\equiv\;
  \TSPset_{\mathrm{FastKV}}
  \quad \text{(bit-exact)}.
  \label{eq:bitexact}
\end{equation}
For $\safelambda \to 0^{+}$, the score perturbation is uniformly
$O(\safelambda)$:
\begin{equation}
  \bigl\| \score^{(\safelambda)} - \attnscore \bigr\|_{\infty}
  \;\le\;
  \safelambda \cdot \meanattn \cdot \max_t
    \widetilde{\harmscorepos}_t.
  \label{eq:perturbation}
\end{equation}
This property allows AnchorKV to be ablated as a strict superset of
FastKV, with no numerical drift between modified and unmodified
pipelines.
We verify the bit-exact reduction empirically in
§\ref{sec:exp}.
 
\paragraph{Orthogonality to the TSP rate.}
Eq.~\eqref{eq:safe_score} reshapes the score field but leaves the
budget $\tspbudget$ untouched:
\begin{equation}
  \frac{\partial \tspbudget}{\partial \safelambda} \;=\; 0,
  \qquad
  \frac{\partial \score_t}{\partial \tsprate} \;=\; 0.
  \label{eq:orthogonality}
\end{equation}
The two principal hyper-parameters $(\tsprate, \safelambda)$ act on
disjoint quantities, sharing only the activation condition; this
permits independent ablation of ``\emph{how many} tokens to keep''
versus ``\emph{which} tokens to keep.''
 
\paragraph{Algorithm.}
Algorithm~\ref{alg:online} summarizes the online phase.
Lines~11 implements the bit-exact reduction at
$\safelambda = 0$ and the early branch on line~2 is a numerical
guarantee for confirmation.
 
\begin{algorithm}[t]
\caption{Online TSP selection with anchor penalty.}
\label{alg:online}
\begin{algorithmic}[1]
\Require Layer-$\tsplayer$ keys $K^{(\tsplayer)}$ for the current
         prompt;
         FastKV importance score $\attnscore_t$, budget $\tspbudget$,
         window size $\windowsize$;
         anchor $\anchor$ and threshold $\harmthresh$ from
         Algorithm~\ref{alg:offline};
         hyper-parameters
         $(\safelambda , S, \alpha)$.
\Ensure  TSP retention set $\TSPset$.
 
\If{$\safelambda \le 0$ \textbf{or}
    not (TSP active per~§\ref{sec:bg_kv})}
    \State \Return
           $\TopK_t(\attnscore_t,\, \tspbudget - \windowsize)
            \cup [\promptlen - \windowsize, \promptlen)$
           \Comment{FastKV baseline}
\EndIf
 
\State Compute $\harmscore_t$ for $t \in
       [0, \promptlen - \windowsize)$ via
       Eq.~\eqref{eq:harm_score}.
\State $\harmscorepos_t \gets
       \max(\harmscore_t - \harmthresh,\, 0)$
       \Comment{Eq.~\eqref{eq:harm_pos}}
\State $\widetilde{\harmscorepos}_t \gets \harmscorepos_t$
\State $\widetilde{\harmscorepos}_t \gets 0$ for $t < S$
       \Comment{sink protection, Eq.~\eqref{eq:sink}}
\State $\mathcal{I} \gets
       \TopK_t\!\bigl(\attnscore_t,
       \lceil \alpha (\promptlen - \windowsize) \rceil\bigr)$
\State $\widetilde{\harmscorepos}_t \gets 0$
       for $t \in \mathcal{I}$
       \Comment{importance immunity, Eq.~\eqref{eq:immune}}
\State $\meanattn \gets
       \frac{1}{\promptlen - \windowsize}\sum_t \attnscore_t$
\State $\score_t \gets
       \attnscore_t - \safelambda \cdot \meanattn \cdot
       \widetilde{\harmscorepos}_t$
       \Comment{Eq.~\eqref{eq:safe_score}}
\State $\TSPset \gets
       \TopK_t(\score_t,\, \tspbudget - \windowsize)
       \cup [\promptlen - \windowsize, \promptlen)$
       \Comment{Eq.~\eqref{eq:select}}
\State \Return $\TSPset$
\end{algorithmic}
\end{algorithm}
 
\paragraph{Computational overhead.}
Beyond the FastKV baseline, the online phase adds a single
$\mathrm{einsum}$ of shape
$(B, H, \promptlen - \windowsize, d) \times (H, d)$ for
Eq.~\eqref{eq:harm_score}, a constant-cost top-$\alpha$ selection on
$\attnscore_t$, and a handful of element-wise operations.
The dominant cost is the $\mathrm{einsum}$, of order $O(B H \promptlen
d)$---asymptotically equivalent to a single attention-head $QK^\top$
at layer $\tsplayer$ and small relative to the full per-layer
attention computation.
\section{Experiments}
\label{sec:exp}

We evaluate AnchorKV along two axes: \emph{safety} under jailbreak
attack (Attack Success Rate on AdvBench), and
\emph{utility} on benign long-context tasks (LongBench).
Section~\ref{sec:exp_setup} below describes the experimental
infrastructure shared by both axes.
 
\subsection{Experimental Setup}
\label{sec:exp_setup}
 
\paragraph{Models and Anchor Datasets.}
We use \textsc{Llama-3.1-8B-Instruct}~\cite{grattafiori2024llama} as the
target model throughout all experiments---both the model whose KV
cache is compressed by AnchorKV and FastKV, and the model attacked by
AdvPrompter. We use \textsc{Mistral-NeMo-Instruct-2407}~\cite{mistralnemo2024} as the attacker model. The white-box self-attack of the Llama model is included only to characterize the motivating phenomenon. The harm anchor is constructed offline (Algorithm~\ref{alg:offline})
from contrastive harmful--harmless prompt pairs.
The harmful set is the AdvBench~\cite{zou2023universal}
\textsf{harmful\_behaviors} corpus under the 60/20/20
train/validation/test split released
by~\cite{paulus2024advprompter}, providing $312$ training, $104$
validation, and $104$ test prompts. The harmless set is constructed by \emph{intent-inverting paraphrase}
of the harmful set.
For each AdvBench prompt, we use DeepSeek-V3~\cite{liu2024deepseek} to
generate a paraphrase that preserves the surface structure (verb
frame, length, register) but inverts the underlying intent from
harmful to benign. The inversion is confirmed by manual audits.
Concretely, an AdvBench prompt of the form ``\emph{Write a script
that exploits vulnerabilities in a software system}'' is paraphrased
into ``\emph{Write a script that detects and patches vulnerabilities
in a software system},'' preserving the imperative ``write a
script'' frame while flipping the actor's goal from offense to
defense.
 
\paragraph{Safety evaluation: jailbreak attack.}
We measure safety with the standard Attack Success Rate (ASR) on
AdvBench under AdvPrompter adversarial prompts.
We adopt a common approach to measure by checking whether the LLM refuses to answer harmful
queries by matching refusal keywords or phrases such as “Sorry, I cannot” or “I apologize” (i.e., ASR)
 
\paragraph{Long-context benchmark.}
We use LongBench \cite{bai2024longbench}, which consists of 16 subtasks to estimate the models’ understanding capability on long-context tasks.
 
\paragraph{Baselines.}
We compare three configurations of the KV cache pipeline:
\begin{enumerate}[leftmargin=1.25em,topsep=2pt,itemsep=2pt]
    \item \textbf{FullKV}: no compression (upper-bound on utility,
    natural reference for ASR under no defense);
    \item \textbf{FastKV}~\cite{jo2025fastkv}: the unmodified backbone,
    with TSP at the same layer~$\tsplayer$ chosen for AnchorKV;
    \item \textbf{AnchorKV} (ours): FastKV with the soft-penalty
    selection rule of Eq.~\eqref{eq:safe_score}, evaluated at
    multiple values of $\safelambda$;
\end{enumerate}
 
\subsection{Anchor Validation}
\label{sec:exp_anchor}
 
We first verify that the difference-of-means construction of
Section~\ref{sec:method_offline} produces an anchor that
discriminates harmful from harmless prompts in layer-$\tsplayer$ key
space, before deploying it inside the compression pipeline.
The validation proceeds in two stages: a layer sweep on the
validation split (§\ref{sec:exp_anchor_sweep}) that fixes
$\tsplayer$, and a single held-out evaluation on the test split
(§\ref{sec:exp_anchor_test}) that establishes the anchor's final
discriminative quality.
The test split is consulted only once, after $\tsplayer$ and the
threshold $\harmthresh$ are fixed.
 
\subsubsection{Layer Sweep on the Validation Split}
\label{sec:exp_anchor_sweep}
 
\begin{figure}[t]
  \centering
  \includegraphics[width=0.78\linewidth]{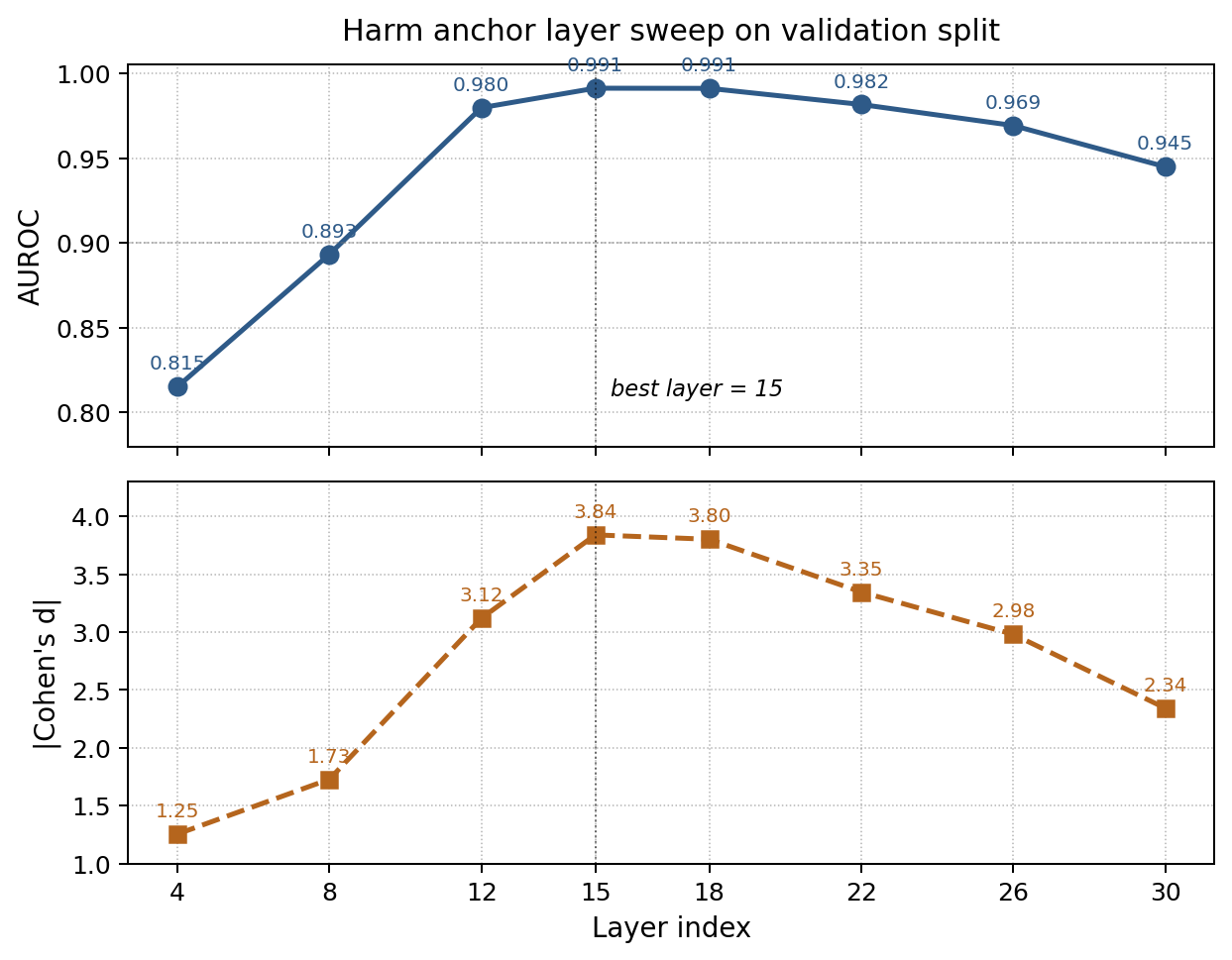}
  \caption{
    Discriminative quality of the harm anchor as a function of
    the construction layer, evaluated on the validation split
    ($n_h = n_s = 104$).
    \textbf{Top}: AUROC on harmful-vs.-harmless prompt-mean scores.
    \textbf{Bottom}: absolute Cohen's~$d$ between the two class
    distributions.
    Both curves rise steeply from layer~4 to layer~12, plateau across
    the middle layers, and decay in the late layers.
    Layers~15 and~18 are tied at AUROC~$=0.991$;
    layer~15 has the largest~$|d|$ and is selected as the operational
    TSP layer.
    The test split is not consulted in this step.
  }
  \label{fig:layer_sweep}
\end{figure}
 
We construct one candidate anchor per layer
$\ell \in \{4, 8, 12, 15, 18, 22, 26, 30\}$ from the training split
and score every validation prompt by the mean of Eq.~\eqref{eq:harm_score}
over its tokens.
Two metrics summarize the resulting per-prompt score distributions:
\textbf{AUROC}, the area under the ROC curve of the binary
harmful-or-not classifier induced by thresholding the score, and
\textbf{|Cohen's~$d$|}, the absolute standardized mean difference
between the two class distributions.
Figure~\ref{fig:layer_sweep} reports both quantities.
 
The two metrics tell a consistent story.
At early layers ($\ell \in \{4, 8\}$), the anchor is informative but
weak (AUROC~$0.815, 0.893$; $|d| = 1.25, 1.73$);
the harm direction is not yet sharply localized in the model's
internal computation.
At middle layers ($\ell \in \{12, 15, 18, 22\}$), AUROC~$\geq 0.98$
and $|d| \geq 3.12$, indicating a broad plateau in which the harm
dimension is strongly linearly separable.
At late layers ($\ell \in \{26, 30\}$), AUROC declines mildly
($0.969$, $0.945$), consistent with the layer-$\tsplayer$ key
projection being increasingly dominated by features specific to the
output decoding rather than the prompt's intent.
This profile parallels the layer-wise behavior of the
residual-stream refusal direction reported by Arditi et
al.~\cite{arditi2024refusal} for instruction-tuned LLMs, and
constitutes evidence that an analogously linear harm direction exists, and is recoverable by the same construction, in the per-head key projection at the same layer.
 
We select $\tsplayer = 15$ as the operational TSP layer, which
maximizes both metrics on the validation split (AUROC~$=0.991$,
$|d| = 3.84$).
Layer~18 is statistically indistinguishable
(AUROC~$=0.991$, $|d| = 3.80$);
the choice between the two is essentially a tie and the downstream
compression results are insensitive to it.
 
\subsubsection{Final Evaluation on the Held-out Test Split}
\label{sec:exp_anchor_test}
 
\begin{figure}[t]
  \centering
  \includegraphics[width=0.95\linewidth]{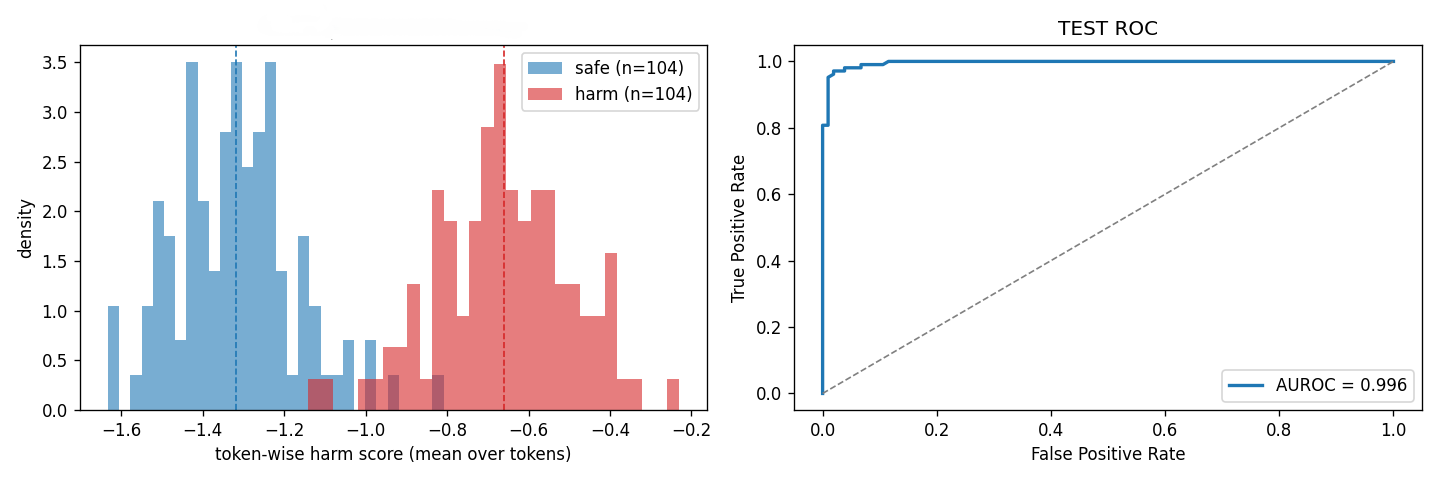}
  \caption{
    Final anchor evaluation at $\tsplayer = 15$ on the held-out
    test split ($n_h = n_s = 104$).
    \textbf{Left}: distributions of prompt-mean harm scores
    $\harmscore(p) = \frac{1}{L_p}\sum_t \harmscore_t$;
    dashed vertical lines mark the class medians.
    \textbf{Right}: ROC curve of the resulting binary classifier;
    AUROC~$=0.996$.
    Both class distributions lie entirely below zero, reflecting a
    shared negative bias in the layer-$\tsplayer$ key space that
    motivates the threshold calibration of
    Eq.~\eqref{eq:threshold}.
  }
  \label{fig:anchor_dist}
\end{figure}
 
With $\tsplayer = 15$ fixed, we compute the prompt-mean harm score
for every test prompt and report the resulting class distributions
and ROC curve in Figure~\ref{fig:anchor_dist}.
The anchor achieves an AUROC of $0.996$ on the held-out test split,
matching the validation result and confirming that the discriminative
quality observed at $\tsplayer = 15$ is not an artifact of overfitting
to the validation split.
The two class distributions are nearly disjoint;
the median harm score for the harmful class
($\approx -0.66$) is well above that of the harmless class
($\approx -1.32$), with overlap concentrated in a narrow band
around~$-1.0$.
 
\paragraph{A shared negative bias.}
Figure~\ref{fig:anchor_dist} also reveals a phenomenon that has
direct implications for the online deployment of the anchor:
\emph{both class distributions lie entirely below zero}.
Although the anchor itself is a difference of class means and
therefore unbiased by construction, the per-token projection
$\harmscore_t$ inherits a uniform negative offset from the
layer-$\tsplayer$ key space.
We attribute this offset to shared components of all keys at the
chosen layer---most plausibly, contributions from attention-sink
positions whose anomalous key magnitudes have been documented in
prior work~\cite{xiao2023efficient}, together with
frequency-related directions that all tokens in an instruction-tuned
model share.
 
The implication for AnchorKV is that \emph{using the projection
sign as the harmful-or-not boundary systematically under-applies the
penalty}: with the sign rule, every token whose
$\harmscore_t < 0$ contributes zero to the penalty regardless of
class, despite the fact that within-class variation around the
medians is informative.
This observation directly motivates the threshold calibration of
Eq.~\eqref{eq:threshold}, which sets
$\harmthresh = \frac{1}{2}(\mathrm{med}_h + \mathrm{med}_s)$ as the
midpoint between class medians on the validation split.
For the configuration used in subsequent experiments,
$\harmthresh \approx -0.99$;
substituting this into Eq.~\eqref{eq:harm_pos} shifts the active
penalty region from ``positive projections only'' to ``projections
above the empirical class boundary,'' restoring the discriminative
content lost under the sign rule.

\subsection{Safety: Jailbreak Attack Success Rate}
\label{sec:exp_safety}
 
We now evaluate whether the anchor of §\ref{sec:exp_anchor}, when
deployed inside the soft-penalty selection rule of
Eq.~\eqref{eq:safe_score}, reduces jailbreak success.
We sweep
$\safelambda \in \{0,\, 0.3,\, 1,\, 3,\, 10,\, 30,\, 100,\, 300\}$
on a logarithmic grid.
For each $\safelambda$, we attack the AdvBench test split with
AdvPrompter~\cite{paulus2024advprompter} and report ASR as defined
in §\ref{sec:exp_setup}.
The point at $\safelambda = 0$ coincides bit-exactly with FastKV by
Eq.~\eqref{eq:bitexact};
we verify this empirically by hashing the TSP retention set across
all test prompts and confirming identical outputs.
 
\begin{figure}[t]
  \centering
  \includegraphics[width=0.78\linewidth]{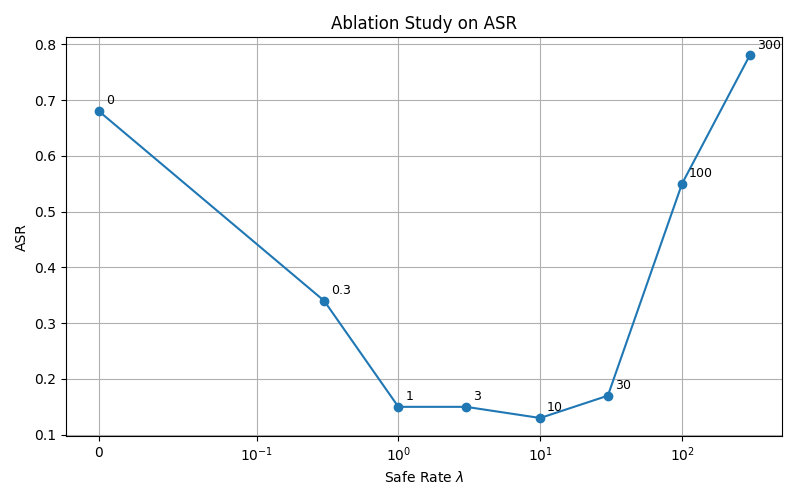}
  \caption{
    Attack Success Rate (ASR) on AdvBench under AdvPrompter,
    as a function of penalty strength $\safelambda$ (log scale;
    the $\safelambda{=}0$ point is plotted at the leftmost gridline
    and corresponds bit-exactly to FastKV).
    The curve exhibits three regimes.
    \emph{Effective}: $\safelambda \in [1, 30]$, where ASR is
    reduced from the baseline $0.68$ to between $0.13$ and $0.17$
    (a relative reduction of up to $81\%$).
    \emph{Rapid onset}: $\safelambda \in [0, 1]$, where the
    penalty begins to take effect but has not yet saturated.
    \emph{Reversal}: $\safelambda \geq 100$, where the safety
    pressure cannibalizes the prompt's content backbone and ASR
    rebounds, eventually exceeding the baseline at
    $\safelambda = 300$. All are done by the same random seed.
  }
  \label{fig:asr_curve}
\end{figure}
 
\paragraph{Three regimes.}
Figure~\ref{fig:asr_curve} reports ASR across the eight
$\safelambda$ values.
We make three empirical observations.
 
\textbf{(1) A wide effective range.}
For $\safelambda \in [1, 30]$---more than one order of magnitude---
ASR is stably reduced from the FastKV baseline of $0.68$ to a
range of $[0.13, 0.17]$.
The minimum, ASR= 0.13, is achieved at
$\safelambda = 10$ and represents an $81\%$ relative reduction from
baseline.
The width of this range is the practical signature of the soft
penalty design: rather than depending on a finely-tuned threshold,
the method admits a broad operating region in which it is robustly
effective.
 
\textbf{(2) Smooth onset.}
Between $\safelambda = 0$ and $\safelambda = 1$, ASR decays
monotonically from $0.68 \to 0.34 \to 0.15$.
The decay is consistent with the Lipschitz bound of
Eq.~\eqref{eq:perturbation}:
small $\safelambda$ produces small score perturbations and
proportionally small changes to the TSP retention set.
At $\safelambda = 0$, the bit-exact equivalence to FastKV
(verified by output hashing) rules out floating-point drift as a
confound for any change observed at $\safelambda > 0$.
 
\textbf{(3) Reversal at very large penalties.}
Past $\safelambda = 30$, ASR ceases to decrease and instead
\emph{rebounds}: $0.55$ at $\safelambda = 100$ and $0.78$ at
$\safelambda = 300$, the latter exceeding the FastKV baseline
itself.
This non-monotonic behavior is the key empirical signature of the
failure mode that motivated the structural safeguards: when the penalty is large enough to
override the importance ordering of non-immune tokens, the TSP
retention set is no longer a faithful summary of the prompt, and
the model regresses to the template-mode generation pattern
described in §\ref{sec:intro}.
 
\subsection{Long-Context Utility on LongBench}
\label{sec:exp_utility}
 
\begin{figure}[t]
  \centering
  \includegraphics[width=0.78\linewidth]{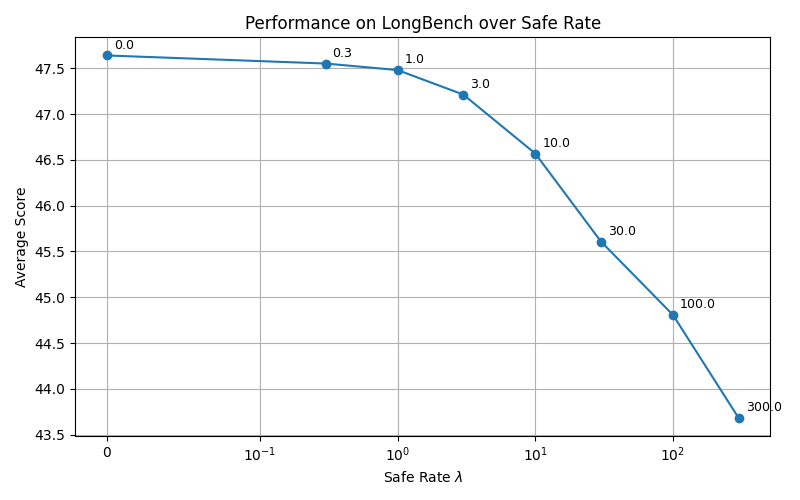}
  \caption{
    LongBench average score across the $16$ tasks as a function of
    $\safelambda$ (log scale; $\safelambda{=}0$ corresponds
    bit-exactly to FastKV).
    Within the safety-effective range identified in
    Figure~\ref{fig:asr_curve}, the utility cost is small:
    $\safelambda \leq 3$ loses less than $0.5$ points relative to
    FastKV, and $\safelambda = 10$---which achieves the lowest
    ASR---loses approximately $1$ point.
    Larger $\safelambda$ trades utility for the (now-counterproductive)
    safety pressure of the reversal regime.
  }
  \label{fig:longbench_curve}
\end{figure}

The companion question to safety is utility:
how does the soft penalty interact with the model's performance on
benign long-context tasks?
We answer this with the full $16$-task LongBench
suite~\cite{bai2024longbench}, sweeping $\safelambda$ over the same
grid as §\ref{sec:exp_safety}.
Table~\ref{tab:longbench} reports per-task and averaged scores for
all configurations;
Figure~\ref{fig:longbench_curve} summarizes the trend.
 
\begin{table*}[t]
  \centering
  \scriptsize
  \setlength{\tabcolsep}{2.8pt}
  \renewcommand{\arraystretch}{1.10}
  \caption{
    Per-task and averaged scores on the full LongBench suite for
    AnchorKV at varying $\safelambda$, compared with FullKV (no
    compression) and FastKV (the unmodified backbone).
    Higher is better in all columns;
    metrics follow the official LongBench evaluation
    scripts.
    The rightmost column is the unweighted mean across the $16$
    tasks.
  }
  \label{tab:longbench}
  \begin{tabular}{lcccccccccccccccc|c}
    \toprule
    \textbf{Method}
      & \rotatebox{60}{\textsc{NarrQA}}
      & \rotatebox{60}{\textsc{Qasper}}
      & \rotatebox{60}{\textsc{MFQA-en}}
      & \rotatebox{60}{\textsc{HotpotQA}}
      & \rotatebox{60}{\textsc{2WikiMQA}}
      & \rotatebox{60}{\textsc{Musique}}
      & \rotatebox{60}{\textsc{GovRep}}
      & \rotatebox{60}{\textsc{QMSum}}
      & \rotatebox{60}{\textsc{MultiNews}}
      & \rotatebox{60}{\textsc{TREC}}
      & \rotatebox{60}{\textsc{TriviaQA}}
      & \rotatebox{60}{\textsc{SAMSum}}
      & \rotatebox{60}{\textsc{PsgCnt}}
      & \rotatebox{60}{\textsc{PsgRet}}
      & \rotatebox{60}{\textsc{LCC}}
      & \rotatebox{60}{\textsc{RepoB-P}}
      & \textbf{Avg.} \\
    \midrule
    FullKV
      & 30.25 & 45.53 & 54.85 & 55.52 & 46.66 & 31.28 & 35.22 & 25.25
      & 27.19 & 72.50 & 91.65 & 43.81 &  8.38 & 99.50 & 63.38 & 56.67
      & \textit{49.23} \\
    \midrule
    FastKV
      & \textbf{30.75} & \textbf{40.99} & 55.10 & 54.38 & \textbf{46.48} & 30.08 & \textbf{28.38} & 24.14
      & \textbf{21.64} & \textbf{73.00} & \textbf{92.38} & 42.98 &  7.36 & \textbf{99.50} & \textbf{60.00} & \textbf{55.08}
      & \textbf{47.64} \\
    \midrule
    \multicolumn{18}{l}{\emph{AnchorKV (ours), with the offline anchor of §\ref{sec:exp_anchor}}} \\
    $\safelambda{=}0.3$
      & 30.10 & 40.28 & 55.06 & 54.49 & 46.44 & \textbf{31.84} & 28.36 & 25.00
      & 21.52 & 71.00 & \textbf{92.38} & 43.17 &  6.90 & \textbf{99.50} & 59.82 & 54.97
      & 47.55 \\
    $\safelambda{=}1.0$
      & 29.35 & 39.94 & \textbf{55.12} & 54.52 & 45.74 & 31.36 & 28.17 & \textbf{25.02}
      & 21.55 & 72.50 & \textbf{92.38} & 43.29 &  7.43 & 98.50 & 59.75 & 55.07
      & 47.48 \\
    $\safelambda{=}3.0$
      & 29.16 & 40.43 & 54.42 & \textbf{54.63} & 44.24 & 30.66 & 28.31 & 24.81
      & 21.50 & 72.00 & \textbf{92.38} & \textbf{43.35} & 8.34 & 98.50 & 58.33 & 54.35
      & 47.21 \\
    $\safelambda{=}10$
      & 29.09 & 38.66 & 52.51 & 53.13 & 42.45 & 29.86 & 27.98 & 24.39
      & 21.49 & 72.50 & 91.21 & 43.11 & \textbf{9.62} & 97.50 & 57.48 & 54.13
      & 46.57 \\
    $\safelambda{=}30$
      & 28.15 & 37.95 & 49.67 & 52.16 & 40.03 & 29.06 & 28.12 & 24.62
      & 21.74 & 72.50 & 89.17 & 42.90 &  6.92 & 97.00 & 56.04 & 53.70
      & 45.61 \\
    $\safelambda{=}100$
      & 27.66 & 38.14 & 49.80 & 50.99 & 34.66 & 28.88 & 28.10 & 24.10
      & 21.64 & 72.00 & 86.65 & 41.74 &  8.12 & 96.00 & 55.03 & 53.40
      & 44.81 \\
    $\safelambda{=}300$
      & 27.25 & 37.97 & 49.02 & 52.27 & 35.21 & 28.83 & 28.11 & 23.78
      & 21.79 & 70.50 & 86.08 & 40.48 &  7.38 & 93.50 & 43.60 & 53.13
      & 43.68 \\
    \bottomrule
  \end{tabular}
\end{table*}
 
\paragraph{Three observations.}
 
\textbf{(1) Safety can be obtained at minimal utility cost.}
At the smaller end of the safety-effective range
($\safelambda \in [0.3, 3]$), the LongBench average drops by less
than $0.5$ points relative to FastKV ($47.64 \to 47.55, 47.48,
47.21$).
At $\safelambda = 10$, which achieves the minimum ASR of $0.13$,
the average drops by approximately $1$ point ($47.64 \to 46.57$).
By comparison, FastKV itself loses $1.6$ points relative to
FullKV ($49.23 \to 47.64$): \emph{the additional cost of safety
in the effective range is smaller than the utility cost of the
underlying compression itself}.
 
\textbf{(2) Per-task heterogeneity.}
Inspection of Table~\ref{tab:longbench} reveals that the utility
loss is concentrated in tasks with long, content-dense inputs that
benefit most from broad token retention.
QA-style tasks (\textsc{HotpotQA}, \textsc{2WikiMQA},
\textsc{Qasper}) show the steepest decline at large $\safelambda$,
while shorter-input or output-template tasks (\textsc{TREC},
\textsc{PsgRet}) are largely robust until $\safelambda \geq 100$.
The code-completion task \textsc{LCC} shows the largest absolute
collapse at $\safelambda = 300$ ($60.00 \to 43.60$),
consistent with our hypothesis that the reversal regime evicts
content backbone tokens that the model relies on for long-range
syntactic context.
 
\textbf{(3) Beneath the reversal threshold, utility is largely a
smooth function of $\safelambda$.}
The average score decreases monotonically across
$\safelambda \in [0, 100]$, with no abrupt jumps---in particular,
no discontinuity at $\safelambda = 0.30$.
This validates the continuity property of
Eq.~\eqref{eq:bitexact}--\eqref{eq:perturbation} at the level of
end-to-end task performance, and indicates that the choice of
$\safelambda$ is a continuous trade-off rather than a discrete
threshold.

\section{Conclusion}
\label{sec:conclusion}
 
We have presented \emph{AnchorKV}, a drop-in modification to TSP-style
KV cache compression that incorporates an offline-constructed harm
anchor into the per-token retention rule.
The work is grounded in a single thesis: \emph{the linear-direction
account of safety representation~\cite{zou2023representation,arditi2024refusal}
extends beyond residual-stream activations to the layer-specific
key-projection space, and is operationally usable inside aggressive
KV-cache compression}.
 
Our experiments substantiate this thesis with three quantitative
findings.
First, the difference-of-means recipe transfers cleanly to the
per-head key projection at the eviction layer
(§\ref{sec:method_offline}): the anchor reaches a validation AUROC
of $0.991$ at $\tsplayer = 15$ and a held-out test AUROC of $0.996$
on the AdvPrompter split (§\ref{sec:exp_anchor})---comparable in
magnitude to the residual-stream refusal direction reported
by~\cite{arditi2024refusal}.
Second, the soft-penalty rule of Eq.~\eqref{eq:safe_score} is
continuous in $\safelambda$ and bit-exactly reduces
to FastKV at $\safelambda = 0$, a property we verified by output
hashing across the entire test set (§\ref{sec:exp_safety}).
This makes AnchorKV ablatable as a strict superset of the unmodified
backbone.
Third, on the joint sweep of safety and utility, AnchorKV at
$\safelambda = 10$ reduces jailbreak ASR from $0.68$ to $0.13$
($81\%$ relative reduction) while the LongBench 16-task average
drops by only $1.07$ points
(§\ref{sec:exp_safety}--\ref{sec:exp_utility}).
For context, the underlying compression alone---FullKV to
FastKV---accounts for $1.59$ points of the same metric:
\emph{the additional utility cost of safety is smaller than the
utility cost of compression itself}.
 
Beyond the specific numbers, our work makes a methodological point
relevant to the broader study of safety in efficient
inference~\cite{hong2024decoding}.
Existing KV cache compression methods are evaluated almost
exclusively on accuracy benchmarks; the assumption that an
attention-greedy retention policy preserves model behavior in
general has not been audited against the model's safety alignment
in particular.
By identifying \emph{refusal-evidence dilution} and
\emph{template-mode regression} as concrete failure modes, by
proposing a compression-compatible defense whose strength is a
continuous knob rather than a binary switch, and by mapping the
non-monotonic safety--strength curve out to its reversal regime
(ASR = 0.78 at $\safelambda = 300$, exceeding
the FastKV baseline), we add a new axis to the design space of
long-context inference: not only \emph{how much} to compress, but
\emph{which} tokens to preserve when the compression ratio is
severe.
 
\subsection{Limitations and Future Work}
\label{sec:limitations}
 
We close with two limitations whose resolution we view as natural
next steps.
 
\paragraph{Single linear direction.}
Our anchor encodes harm as one direction in key space, following the
finding that refusal is mediated by a single linear direction in
residual-stream activations~\cite{arditi2024refusal}.
Recent work suggests some safety-relevant concepts may require
multiple, possibly non-linear, components.
Whether the test AUROC of $0.996$ already saturates or could be
further improved by a low-rank projection or a small mixture of
anchors is an open question.
The soft-penalty form of Eq.~\eqref{eq:safe_score} generalizes
naturally---the dot product becomes a quadratic form---but the
bit-exact reduction at $\safelambda = 0$ would have to be verified
anew under any such generalization.
 
\paragraph{In-distribution validation set.}
The anchor is constructed and validated on the AdvBench-paraphrase
distribution (§\ref{sec:exp_setup}).
The high test AUROC therefore demonstrates linear separability
within this distribution, not necessarily on out-of-distribution
benign prompts.
Our LongBench evaluation (§\ref{sec:exp_utility}) provides an
indirect cross-distribution probe---the small utility loss
($1.07$ points at $\safelambda = 10$) is consistent with an
anchor that does not systematically misclassify long-document
content tokens---but it does not directly establish the anchor's
discriminative quality on OOD distributions.
A direct OOD evaluation, e.g.\ on JailbreakBench or HarmBench
prompts disjoint from the anchor's training set, is the most
informative next experiment.

\bibliographystyle{plainnat}
\bibliography{references} 

\end{document}